\documentclass[10pt,twocolumn,letterpaper]{article}

\usepackage[breaklinks=true,bookmarks=false]{hyperref}
\usepackage{iccv}
\usepackage{times}
\usepackage{epsfig}
\usepackage{graphicx}
\usepackage{amsmath}
\usepackage{amssymb}

\iccvfinalcopy 

\def\iccvPaperID{****} 

\begin{document}

\title{Learnable Exposure Fusion for Dynamic Scenes}

\author{Fahd Bouzaraa\\
Huawei German Research Center\\
{\tt\small bouzaraa.fahd@huawei.com}
\and
Ibrahim Halfaoui\\
Technical University of Munich\\
{\tt\small Ibrahim.halfaoui@tum.de}
\and
Onay Urfalioglu\\
Huawei German Research Center\\
{\tt\small onay.urfalioglu@huawei.com}
}

\maketitle

\begin{abstract}
In this paper, we focus on Exposure Fusion (EF)~\cite{ExposFusi2} for dynamic scenes. The task is to fuse multiple images obtained by 
exposure bracketing to create an image which comprises a high level of details. 
Typically, such images are not possible to obtain 
directly from a camera due to hardware limitations, e.g., a limited dynamic range of the sensor.
A major problem of such tasks is that the images may not be spatially aligned due to scene motion or camera motion.
It is known that the required alignment by image registration problems is ill-posed. 
In this case, the images to be aligned vary in their intensity range, which makes the problem even more difficult.

To address the mentioned problems, we propose an end-to-end \emph{Convolutional Neural Network} (CNN) based approach to learn to estimate 
exposure fusion from $2$ and $3$ Low Dynamic Range (LDR) images depicting different scene contents. 
To the best of our knowledge, no efficient and robust CNN-based end-to-end approach can be found in the literature for this kind of problem. The idea is to create a dataset with perfectly aligned LDR images to obtain ground-truth
exposure fusion images. At the same time, we obtain additional LDR images with some motion, 
having the same exposure fusion ground-truth as the perfectly aligned LDR images. 
This way, we can train an end-to-end CNN having misaligned LDR input images, but with a proper ground truth exposure fusion image.
We propose a specific CNN-architecture to solve this problem. In various experiments, we show that the proposed approach yields excellent results.

\end{abstract}

\section{Introduction}

High Dynamic Range imaging (HDRI) emerged in recent years as a major research topic in the context of computational photography, where the main purpose is to bridge the gap between the dynamic range native to the scene and the relatively limited dynamic range of the camera. As a result, the level of details in the captured LDR image is significantly enhanced, so that the final image presents a balanced contrast and saturation in all parts of the scene. The most common approach to render an HDR image with a camera presenting a limited dynamic range is called \emph{Exposure Bracketing}~\cite{RadioCalib2, AcqusitionPipe1, dynamicRange99imp}. It relies on merging several LDR images of the same scene captured with different exposures. By alternating the exposure settings between under-exposure and over-exposure, the input stack of LDR images contains various sets of details in different areas of the scene. These details are combined together into a single HDR image by estimating the inverse of the \emph{Camera Response Function} (CRF). The resulting image finally undergoes a tone-mapping operation for display purposes.

 Another alternative is called \emph{Exposure Fusion} (EF)~\cite{ExposFusi2}. The main difference between both approaches is that exposure fusion directly merges the input LDR images to produce a final high-quality LDR image without the usage of the CRF. The visual characteristics of the resulting fused image are similar to a tone-mapped HDR image (pseudo-HDR image). The direct merging of the input images represents a clear advantage over exposure bracketing since prior information about the exposure settings are not needed and no estimation of the inverse CRF is required. This results in a decrease of the computational complexity while yielding high quality enhancement results.
	
 Both exposure bracketing and exposure fusion are based on the assumption that the input LDR images are aligned. However, misalignment due to camera- of scene-motion will almost always occur, especially when the input images are captured sequentially. As a result, the output image contains strong artifacts where several instances of the same object can be seen. These artifacts are known as the \emph{Ghost Effect}. Whether the HDRI system is based on exposure bracketing or exposure fusion, removing these artifacts from the final image is a very challenging task.
 
  In this work, we aim at taking advantage of the latest advances achieved by ConvNets in classification and image enhancement topics. In a nutshell, our main goal is to combine the tasks of details enhancement and the removal of motion-induced ghost artifacts into a single framework. This is achieved by creating an \textbf{end-to-end mapping which learns the exposure fusion for dynamic scenes}. In other words, our trained model yields a final artifact-free image which disposes of a wider range of details, based on input LDR images presenting motion-related scene differences and color/exposure differences. Similar to exposure fusion, the output of our trained model is a LDR image, as no true HDR transformation is occurring. However, the visual attributes of the resulting image allows it to be labeled as a pseudo-HDR image.
 
  We test our learnable exposure fusion approach for dynamic scenes on several indoor and outdoor scenes and we show that the quality of our results improves upon state-of-the-art approaches. We show as well that our approach is capable of handling extreme cases in terms of motion and exposure difference between the input images, while maintaining a very low execution time. This makes it suitable for low-end capturing devices such as smartphones.

\section{Related Work}
Rendering an artifact-free HDR image or pseudo-HDR image of a dynamic scene is a thoroughly investigated topic. Several approaches claim to successfully handle the misalignment and the associated inconsistencies, so that the final image is ghost- and blur-free. The sphere of these methods can be split into two major categories. 

 The first category falls under the scope of the \emph{De-ghosting} methods. The idea behind these approaches is to select a LDR image from the input stack and use it as a reference in order to detect inconsistencies caused by dynamic pixels in the non-reference images. The subsequent merging procedure aims at discarding the detected inconsistencies from the resulting image. De-ghosting approaches are the methods of choice in scenarios where the computational cost of the enabling algorithm needs to be low. Nonetheless, scenes with large exposure and scene differences might be very challenging for these methods. Motion-related artifacts can still be seen in case of non-rigid motion or large perspective differences in comparison to the reference LDR image. A detailed examination of these methods is provided in~\cite{tursun2015state}.
   
 The second category is composed of approaches relying on sparse and/or dense pixel correspondences in order to align the input images. The alignment can be either spatial where the non-reference LDR images are warped to the view of the selected reference image, or color-based by aligning the reference LDR image to each non-reference LDR image separately using color mapping. In both cases, the goal is to form a stack of aligned but differently exposed LDR images corresponding to the view of the reference image.

 In \cite{kang2003high}, Kang \emph{et al.} introduced an approach which uses optical flow in order to align the input differently exposed images, in the context of video HDRI. Likewise, Zimmer \emph{et al.} propose in~\cite{zimmer2011freehand} a joint framework for \emph{Super-Resolution} and HDRI by aligning all images to the reference view using optical flow. The described approach gets around the issue of color inconsistency for optical flow by including a gradient constancy assumption in the data term of the energy function. Alternatively, Sen \emph{et al.} describe in~\cite{PatchBasedHDR} a solution for simultaneous HDR image reconstruction and alignment of the input images using a joint patch-based minimization framework. The alignment is based on a modified version of the \emph{PatchMatch} (PM)~\cite{Barnes2010Patch} algorithm. The final HDR image is rendered from the well-exposed regions of the reference LDR image and from the remaining stack of LDR images for low-exposed regions in the reference. Likewise, Hu \emph{et al.} propose in~\cite{Hu2013Hdrdeg} to align every non-reference LDR image to the selected reference image, which typically has the highest number of well-exposed pixels. The patch-based alignment approach uses the generalized PM algorithm for well-exposed patches in the reference LDR image and suggests an additional modification to PM for over- or under-exposed patches in the reference image. The final HDR image is composed using the \emph{Exposure Fusion} algorithm. More recently, Gallo \emph{el al.} proposed in~\cite{gallo2015locally} an approach based on the matching of sparse feature points between the designated reference and non-reference images. The \emph{matcher} developed for this purpose is robust towards saturation. Once a dense flow field is interpolated, the warped images and the reference LDR image are merged using a modified exposure fusion algorithm, which minimizes the effects of faulty alignment.

 These methods usually achieve accurate alignment results, which in turn helps creating an artifact-free final HDR or pseudo-HDR image. However, the main limitation of these approaches is related to the high computational cost, which hinders their deployment on devices with limited computational resources such as smartphones. In addition, smaller stacks of input LDR images with significant exposure and scene differences due to large motion, hampers the generation of an artifact-free final image.    
 
\subsection{Convolutional Neural Networks}

Recently, \emph{Convolutional Neural Networks}~\cite{long2015fully} were successfully deployed to low-level image processing applications such as \emph{Image Denoising}~\cite{Viren09Natural,karbasi2013iterative} or \emph{Image Super Resolution}~\cite{dong2014image}. The notable quality enhancement brought by CNNs to these applications explains our interest in developing a fast and robust learnable exposure fusion for challenging dynamic scenes. 

 Considering our target application, taking the global input properties into account is fundamental. The common CNN approach consisting of applying a sliding kernel window for single pixel prediction is expected to be computationally demanding and limited in terms of accounting for the global properties of the input images. Alternative CNN architectures have been recently proposed. Among these, the \emph{FlowNet} architecture which was introduced by Dosovitskiy \emph{et al.} in~\cite{Dobrovsky15Flown} for motion vectors estimation seems to be well-suited to our learnable exposure fusion application. The concept of \emph{FlowNet} is based on a two-stage architecture with a \emph{contractive} part and a following \emph{refinement} part. Both parts are connected by means of long-range links. The contractive side of the network is composed of a succession of convolutional layers with a gradually decreasing spatial resolution of the feature maps. Accordingly, the refinement part contains a sequence of de-convolutional layers with gradual increase of the corresponding spatial resolution.
 
 The contractive part of the network is responsible for extracting high-level features and spatially down-sizing the feature maps. This in turn enables the effective aggregation of information over large areas of the input images. However, the output feature maps at the bottom of this part have a low spatial resolution. Consequently, the role of the refinement part is to simulate a coarse-to-fine reconstruction of the downsized representations by gradually up-sampling the feature maps and concatenating them with size-matching feature maps from the contractive part. This allows for a more reliable recovery of the details lost on the contractive side of the network. The final output is typically a dense per-pixel representation with the same resolution as the input images.

 In the next sections, we first experiment with a basic \emph{FlowNet} architecture for the purpose of learning exposure fusion for dynamic scenes. Next, based on the analysis of the results from the \emph{FlowNet}-based results, we propose a more elaborate architecture which fits the requirements of our application.
 
\section{FlowNet-based Experiments}
\subsection{Dataset}

 The set of images used to train our learnable exposure fusion model typically consists of several scenes. Each scene comprises differently-exposed LDR images depicting various scene contents due to motion, together with the corresponding artifact-free exposure fusion image, which is rendered from aligned but differently-exposed instances of the selected reference LDR image. However, capturing differently exposed but aligned LDR images in a sequential manner is a challenging task, as no motion can be tolerated. We found that we can circumvent this issue by relying on stereo datasets. In fact, the stereo setup provides us with the required configuration for our training set, as one camera (left or right) can be set as the reference view and hence used to obtain the \textbf{aligned but differently exposed} input images as well as the reference LDR image. On the other hand, the second camera provides the needed motion as a result of the spatial shift between both capturing devices. Consequently, the images captured using the second camera are differently exposed and depict a different scene content in comparison to the selected reference image from the first camera.
 
  In our work, our training set is a combination $2$ different datasets. The $1^{st}$ dataset is based of the $2005$, $2006$ and $2014$ \emph{Middlebury Stereo-sets} proposed by Scharstein and Pal in~\cite{scharstein2007learning} and Scharstein \emph{et al.} in ~\cite{schars2014high}. The $2005$ and $2006$ sets contain several scenes composed of $3$ differently exposed LDR images of the left view as well as $3$ additional exposures of the right view. The $2014$ set offers $6$ different exposures of each view. The \emph{Middlebury} datasets enclose challenging scenes, especially in terms of exposure differences and saturated images, but lacks the required scenes diversity as they are captured in an indoor controlled environment. In order to compensate for the lack of outdoor scenes, we create a second stereo-based dataset using $2$ \emph{IDS uEye} cameras with identical settings. The complementary second dataset we captured is composed of several outdoor scenes, each containing $5$ LDR images of the left view and $5$ additional LDR images of the right view. Therefore we are able to train our model on indoor and outdoor scenes simultaneously. Additionally, we apply data augmentation consisting of flipping each image from the training set along the vertical, horizontal and diagonal axes in order to increase the size of training set.
  
 For our initial testings, we limit the number of input images to a pair of under-exposed and over-exposed LDR images. For all pairs of inputs in the training set we set the under-exposed LDR image from the left camera as the corresponding reference input image, and the over-exposed LDR image from the right camera as the non-reference input image. This means that the ground-truth exposure fused image used for training in each sequence is gained using the reference image and the \textbf{over-exposed image from the left camera}. Moreover, we only select LDR image pairs where the exposure ratio between the left dark and right bright images is at least $8$. This guarantees that the trained model is capable of handling input images with large exposure ratios. This results in $3080$ training pairs of input LDR images together with the ground-truth exposure fused image of the reference view. Additionally, $352$ pairs of dark and bright LDR images are available for validation purposes

\subsection{FlowNet-Based Tests}
We experiment in this section with a basic \emph{FlowNet} architecture composed of $3$ convolutional layers in the contractive part and $3$ deconvolutional layers in the refinement part. We use the \emph{Caffe} framework~\cite{jia2014caffe} for the implementation of the layers in the \emph{FlowNet} network. Concerning the network parameters, we set the filter number for all convolutional and deconvolutional layers to $16$, with a $4 \times 4$ convolutional (and deconvolutional) kernel. The stride $s$ for all convolutions and deconvolutions is set to $2$. This enables the down-sampling (or up-sampling) of the feature maps, acting therefore as a pooling layer. The input LDR images are spatially resized to $800 \times 480$ and presented to the network as a concatenated input tensor.  Furthermore, we set the learning rate to $10^{-2}$ which decreases according to a polynomial decay scheme with power equal to $0.9$. The momentum is set to $0.9$. The training and testing of our models are conducted on a NVIDIA TITAN X.
\begin{figure}[h!]
     \includegraphics[scale=0.445]{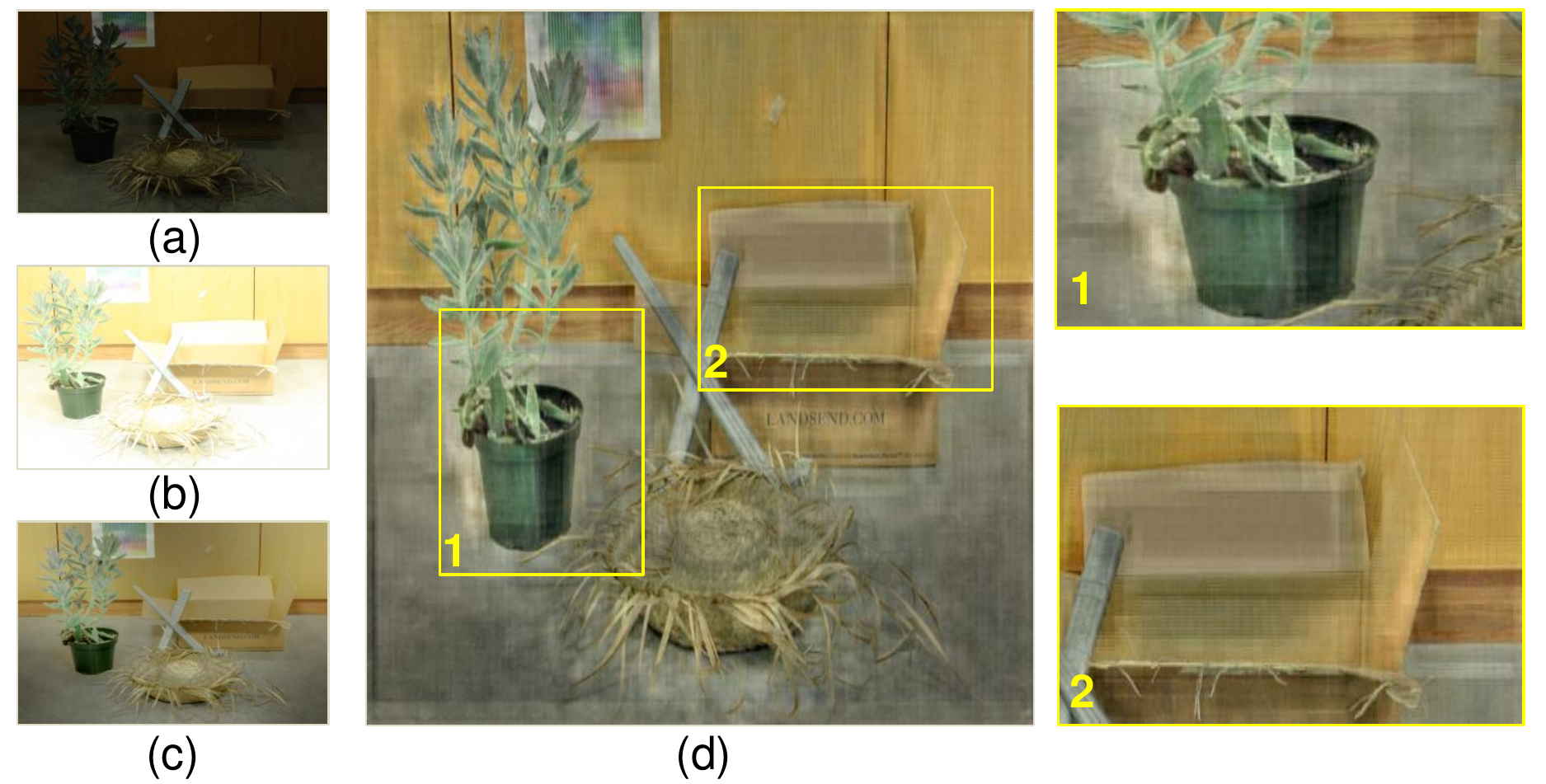}
    \caption{Learnable exposure fusion results (d) on input LDR images (a,b) from the validation set using the \emph{FlowNet} architecture, alongside the corresponding ground-truth exposure fusion image (c) of the reference view (a). Although The rendered CNN-based image presents an improved level of details, artifacts in many areas in the image are visible, as shown in the highlighted regions. Input images courtesy of Daniel Scharstein.}
    \label{fig:Fig3}
\end{figure}
 An example of a gained output image from a pair of LDR images belonging to the validation set is shown in Fig.~\ref{fig:Fig3}. Although the trained model significantly expands the range of depicted details in comparison to the input reference LDR image, clear square-shaped artifacts can be seen. These artifacts are explained by the fact that the refinement stage of the network is unable to reconstruct the image details lost due to down-sampling in the contractive part. In addition, the trained model is conflicted between learning to improve the representation of details in all regions of the reference image, and learning to suppress the inconsistencies between the input images due to the motion/scene difference. Finally, the ground-truth exposure fusion images used for training our model might constraints a wider expansion of the range of presented details. This is particularly observed when the exposure ratio between the input LDR images used to create the ground-truth exposure image, is very high. In such cases, the exposure fusion algorithm used to create our ground-truth image produces visible artifacts in areas which are simultaneously under-exposed and over-exposed in the input images.
 
 Considering all these observations, we propose several modifications to the \emph{FlowNet} architecture used previously. The main goals of the modifications are:
\begin{itemize}
\item Reduce the square-shaped artifacts in the output image by improving the connection between the contractive and refinement parts of the network.
\item Propose an alternative formulation for the task of learnable exposure fusion for dynamic scenes, by breaking it down to several sub-problems that are easier to model. 
\item Ensure a high image quality for the ground-truth exposure fusion images in the training set. The goal here is to increase the level of depicted details in the ground-truth images for all possible scenarios including challenging cases in terms of motion and color differences.
\item Integrate available priors such as the exposure fusion images created directly from the input LDR images. Although these images contain ghost-artifacts, they present valuable priors to our model. In the following, we will refer to these images as \emph{ghost-fused images}.
\end{itemize}

\section{Proposed Modifications}
\subsection{Reducing Reconstruction Artifacts} 

One of the limitations noticed on the results of the basic \emph{FlowNet} architecture is to the loss of details in the contractive part, which hindering their accurate reconstruction during the refinement stage of the network. To tackle this issue, we propose to modify the basic \emph{FlowNet} architecture as suggested in~\cite{Ibrahim2016Segm}. 

 Similar to the basic \emph{FlowNet}, the network architecture proposed in~\cite{Ibrahim2016Segm} is composed of a contractive part and a refinement part. However, the difference to the original \emph{FlowNet} lies in the additional long-range links (concatenations) which connect both parts. The added inputs enforce the redundancy of inherent information before each layer, which in turn enables a better recovery of the details at the output. At each convolutional or deconvolutional layer, the feature maps representing different high-level abstractions from previous layers are combined (concatenated) after accordingly adjusting the resolutions into a single input chunk, as illustrated in. The output of each layer is either convolved and/or deconvolved using the corresponding stride in order to match the dimensions of the target layers. The resulting feature maps are then concatenated to the input of the corresponding later layer.
 
\subsection{Simplifying the Task Formulation}
\begin{figure*}[h!]
     \includegraphics[width=\textwidth, height = 58mm]{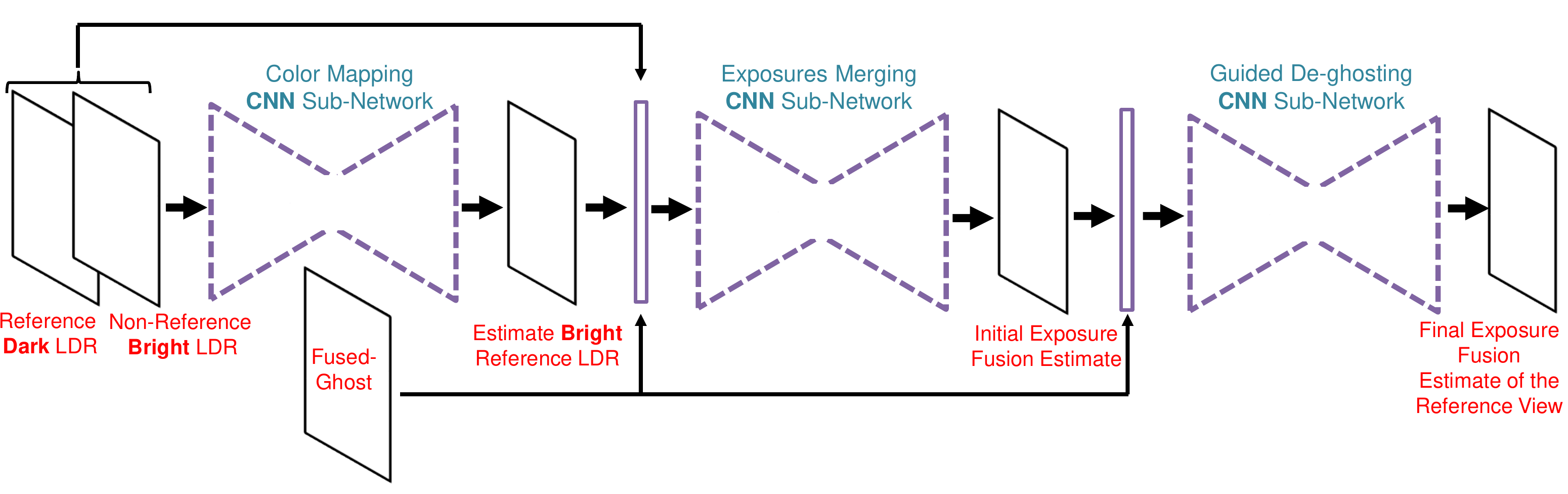}
    \caption{Representation of the proposed architecture composed of the \textbf{color mapping}, \textbf{exposures merging} and \textbf{guided de-ghosting} sub-networks.}
    \label{fig:Fig5}
\end{figure*}
 Numerous state-of-the-art approaches~\cite{Hu2013Hdrdeg,PatchBasedHDR,hu2012exposure} perform several pre-processing operations on the input images prior to the actual HDR/pseudo-HDR merging step. These operations aim at aligning the input images to the selected reference view. By analogy, we propose to split the CNN-based rendering task into three main sub-problems: \textbf{color mapping}, \textbf{exposures merging} and \textbf{guided de-ghosting}. Each sub-problem is represented through a \emph{FlowNet}-inspired sub-network with the modifications proposed in~\cite{Ibrahim2016Segm}. Accordingly, these sub-networks are connected together so that they form the desired end-to-end mapping between the input pair of LDR images and the output image of the reference view. An illustration of the proposed configuration is shown in Fig.~\ref{fig:Fig5}. 
  
  The first convolutional sub-network learns the color mapping model between the differently exposed input LDR images. More specifically, this step aims at estimating the over-exposed instance of the reference under-exposed LDR image. Training such a model is possible since our dataset contains the differently exposed instances of each view, which we originally used to create the ground-truth exposure fusion images. 
  
   Next, the estimate of the over-exposed version of the reference LDR image is forwarded to the subsequent exposures merging sub-network. In theory, the reference LDR image and the output of the previous color mapping sub-network are the only images required for generating the output (pseudo-HDR) image of the reference view. However, our tests have shown that providing the input non-reference image significantly enhances the quality of the output image, as it contains scene details which might not be present in the reference LDR image or in the output of the color mapping stage. On the other hand the perspective shift in comparison to the reference image causes no visible artifacts in the final image. Furthermore, we provide the so-called \emph{ghost-fused} image as an additional input to the second sub-network. As mentioned earlier, the ghost-fused image is gained from the input LDR images using the exposure fusion algorithm and contains ghost artifacts due to the difference in terms of scene content between the input images.
  
  The final \emph{guided de-ghosting} sub-network enables the enhancement of the previously estimated pseudo-HDR image, using to this end the ghost-fused image as an additional input. Thus the input to the third sub-network is composed of the ghost-fused image and the initial output (pseudo-HDR) image estimate from the previous sub-network. The final output image of the guided de-ghosting sub-network contains more details and hence a wider dynamic range than the output of the second sub-network estimate. 

\subsection{Improving the Quality of the Ground-Truth Images}
\begin{figure}[h!]
     \includegraphics[scale=0.46]{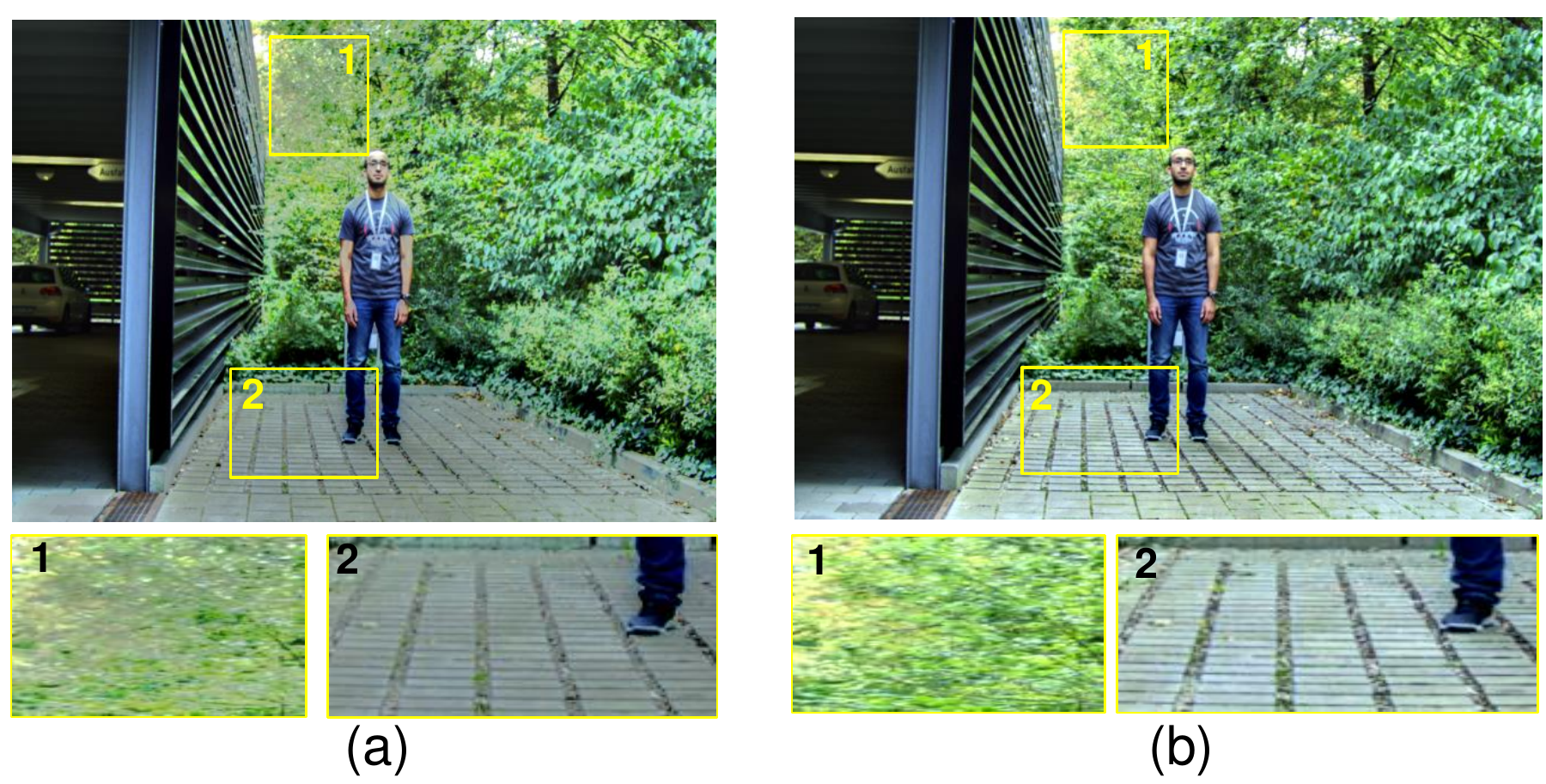}
    \caption{Illustration of the difference between the $2$-LDR-based ground-truth image \textbf{(a)} and the $5$-LDR-based ground-truth image \textbf{(b)}. Clearly, the $5$-LDR-based image is more suitable to train the desired exposure fusion for dynamic scenes model.}
    \label{fig:Fig6}
\end{figure}

As mentioned previously, the generation of the ground-truth exposure fusion images used for training is based on the exposure fusion algorithm. Apart from its relative straightforwardness and performance stability, exposure fusion does not require any priors on the input stack of LDR images, such as the corresponding exposure times or ratio. However, in the case of $2$ input LDR images with large exposure difference, the resulting output image contains visible artifacts as shown in Fig.~\ref{fig:Fig6}.

 Therefore, training on low-quality ground-truth images impacts negatively the performance of the model in terms of details expansion. To solve this problem, we propose to use ground-truth exposure fusion images composed from the \textbf{full stack} of available LDR images. For example, in the case of the dataset we created using the uEye cameras, the ground-truth exposure fusion image for each scene is gained from the merging of the $5$ differently exposed instances of the dark (under-exposed) view. This way, our model does not only render a ghost-free visually enhanced image, but also simulates the case where more than $2$ input LDR images are available. This allows to deal with very challenging cases in terms of exposure differences and number of input images. Fig.~\ref{fig:Fig6} shows the quality difference between the $2$-LDR-based ground-truth image and the $5$-LDR-based ground-truth image.

\section{Experiments}
  
 Taking into account the proposed modifications on the basic \emph{FlowNet} architecture, we train an exposure fusion for dynamic scenes model according to the architecture presented in Fig.~\ref{fig:Fig5}. The color mapping sub-network is composed of $5$ convolutional layers and $5$ deconvolutional layers using the design modifications proposed in~\cite{Ibrahim2016Segm}, whereas the remaining sub-networks are composed of $3$ convolutional layers and $3$ deconvolutional layers (also according to the design of ~\cite{Ibrahim2016Segm}). We set the filter numbers of all layers to $32$ for the color mapping sub-network and reduce this number to $16$ for the remaining sub-networks.The network configuration is similar to the initial tests made using the basic \emph{FlowNet} architecture.

\begin{figure*}[t]
     \includegraphics[width=\textwidth, height = 145mm]{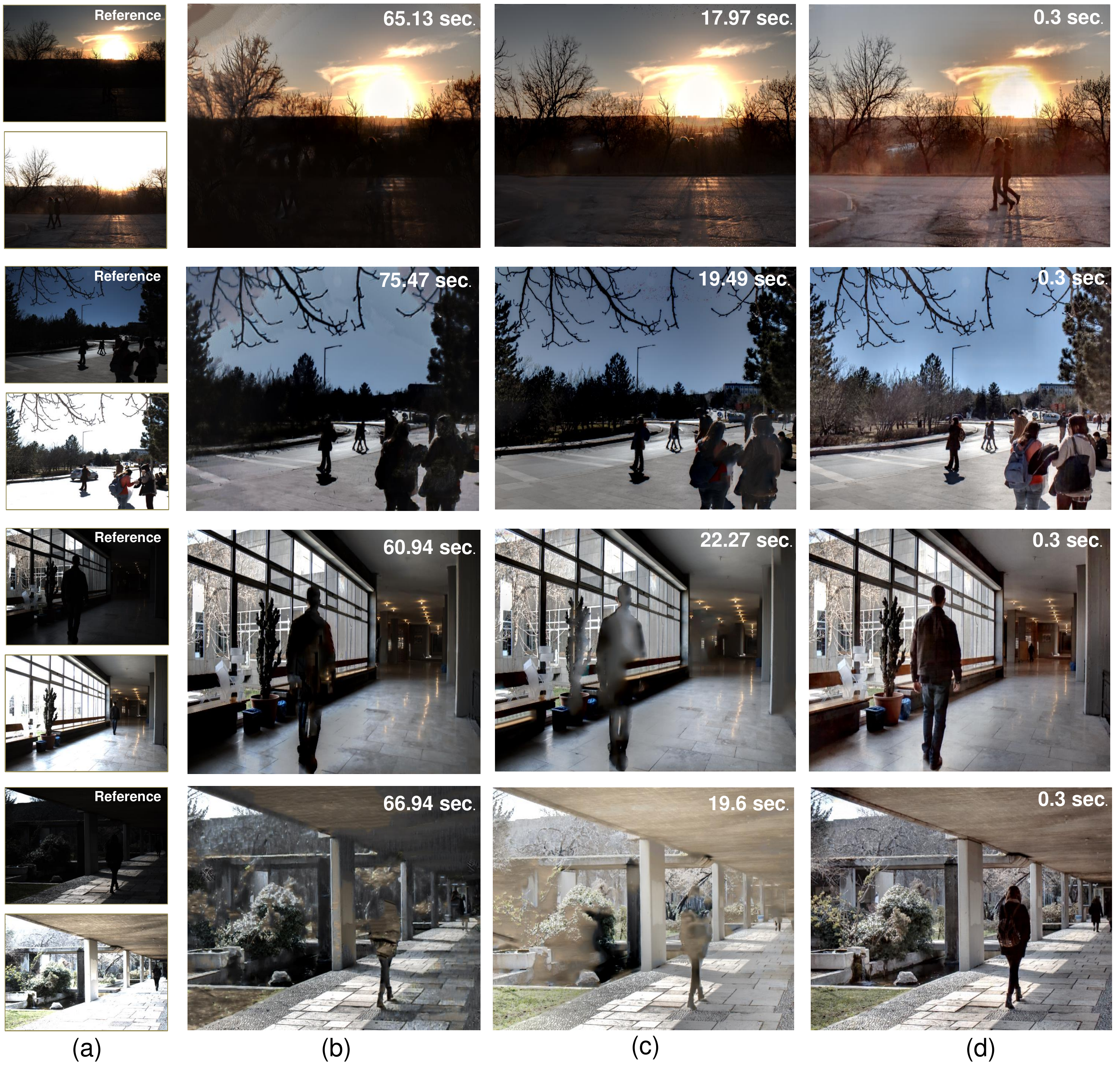}
    \caption{Visual comparison between the methods of Sen \emph{et al.} (b), Hu \emph{et al.} (c) and our results (d), together with the corresponding execution times (only for the alignment part in the case of Sen \emph{et al.} and Hu \emph{et al.}. Note that we used exposure fusion to generate the corresponding result images for Sen \emph{et al.} and Hu \emph{et al.}. Clearly, our results are free from artifacts and yield the highest expansion of the range of details, despite the challenging nature of the scene in terms of exposure times difference between the input LDR images (a) and the depicted motion. Input images courtesy of Okan Tarhan Tursun~\cite{tursun2016objective}.}
    \label{fig:Fig8}
\end{figure*}

  Figure~\ref{fig:Fig8} contains a comparison between the methods of Sen \emph{et al.}~\cite{PatchBasedHDR}, Hu \emph{et al.}~\cite{Hu2013Hdrdeg} and our proposed model. The methods of Sen \emph{et al.} and Hu \emph{et al.} propose a simiar framework as ours, namely by initially aligning the input non-reference LDR images to the view of the reference image. For these comparisons, we used the MATLAB implementations provided by the authors. Based on these implementations, the input LDR images are aligned colorwise to the designated reference LDR image and merged subsequently into the final pseudo-HDR image using exposure fusion. We sought as well to compare our results to the method of Gallo \emph{et al.} introduced in~\cite{gallo2015locally}, but no source code/executable of the mentioned approach were available.
  
 The scenes presented in Fig.~\ref{fig:Fig8} neither belong to the training nor to the validation sets. Various types of motion are also represented in these images such as object and/or scene motion. In addition, the ground-truth exposure fusion images for these scenes are not available, as the aligned and differently exposed instances of the reference LDR images are not provided~\cite{tursun2016objective}. Despite the large exposure and perspective differences between the input LDR images, our model successfully extends the dynamic range of the reference LDR image with no artifacts related to moving objects in the scenes. This is however not the case for the results of Sen \emph{et al.} and Hu \emph{et al.}, where the image alignment operation based on PatchMatch fails to track dynamic objects especially in the over- or under-exposed areas. This results in clear artifacts as shown in Fig.~\ref{fig:Fig8}. Furthermore, the improved details representation of our results come with low execution times in comparison to the other methods. All experiments were conducted on a computer with standard configuration. 
  
\begin{figure*}[h!]
     \includegraphics[width=\textwidth, height = 75mm]{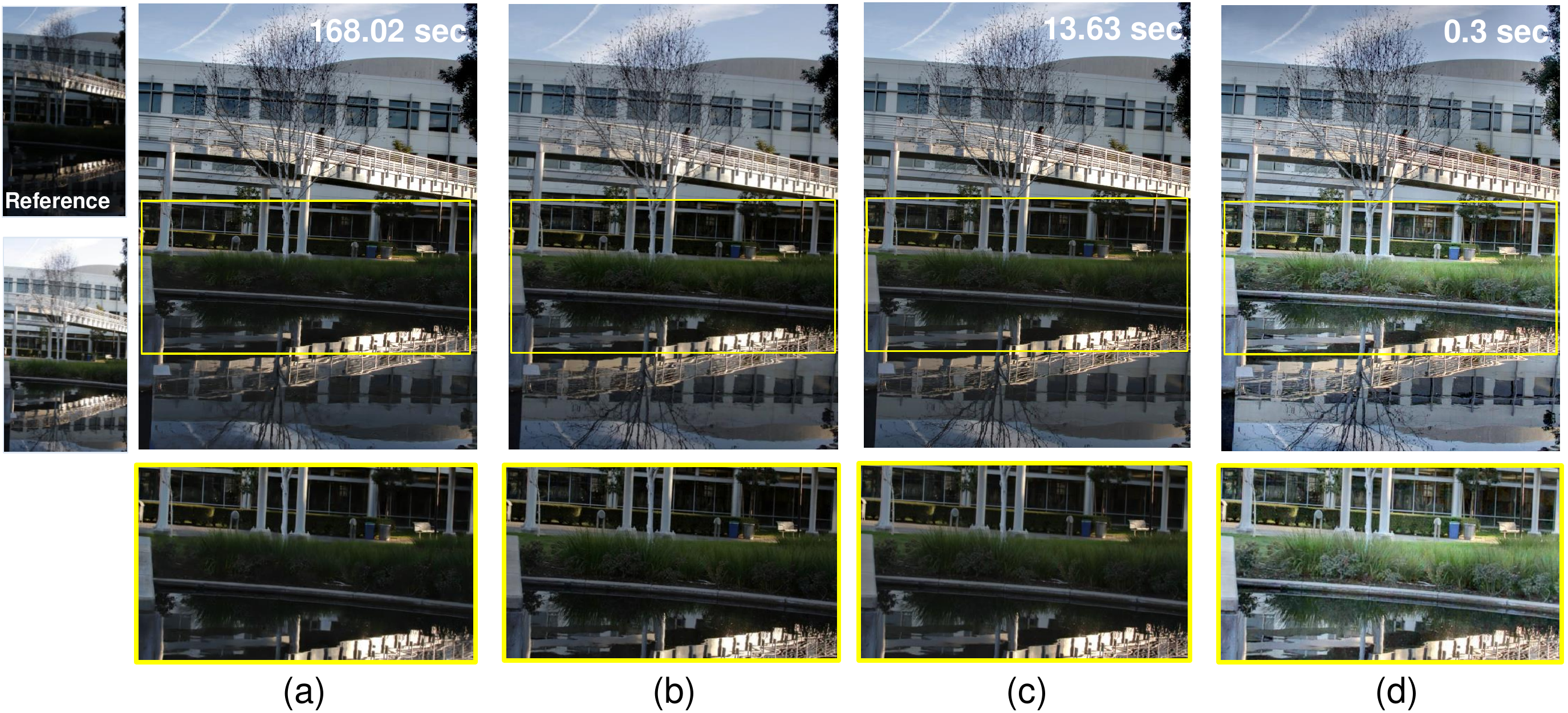}
    \caption{Visual comparison between the methods of Sen \emph{et al.}~\cite{PatchBasedHDR} (a), Hu \emph{et al.} (2012)~\cite{hu2012exposure} (b), Hu \emph{et al.} (2013)~\cite{Hu2013Hdrdeg} (c) and our results (d), together with the corresponding execution times. Note that we used exposure fusion to generate the results for Sen \emph{et al.} and Hu \emph{et al.} (2013). Our rendering model trained on ground-truth exposure fusion images images gained from an extended stack of input LDR images significantly extends the dynamic range of the reference LDR image, so that details in all possible areas are visible and most importantly in regions which are under- and/or over-exposed in the input images (highlighted areas). Images courtesy of Jun Hu~\cite{hu2012exposure}.}
    \label{fig:Fig9}
\end{figure*}

  Moreover, our results exhibit an extended dynamic range, far beyond the dynamic range available in the input LDR images. This can be also observed on the challenging scene shown in Fig.~\ref{fig:Fig9}. Our output image presents a well balanced quality in terms of details and color representation. This is particularly notable in the indoor part of the scene (see the highlighted areas in Fig.~\ref{fig:Fig9}), where the comparison shows that our model is the only method capable of retrieving and depicting the details in this area. Accordingly, the extended dynamic range is a result of the above introduced modifications, where we train our model on ground-truth exposure fusion images merged from the full stack of available LDR images. 

  Additionally, we tested the performance of our proposed model on scenarios where the input stack consists of $3$ LDR images. In such cases, the input stack is composed of an under-exposed (dark) image, a mid-exposed image and an over-exposed (bright) image. The mid-exposed LDR image is designated as the corresponding reference view, as it contains more scene details in comparison to the under- and over-exposed images. This implies that in our case, the network configuration discussed earlier needs to be modified since $2$ color mapping sub-networks are now required in order to obtain the estimates of the under-exposed as well as the over-exposed instances of the reference image. 
  
  Accordingly, the $3$-LDR-based rendering scenario imposes different constraints on the training set as well as on the CNN architecture of each sub-network. Consequently, we reshuffle the training set used for the $2$-LDR-based cases by changing the configuration of each scene so that the reference LDR image is \emph{mid-exposed} in comparison to the non-reference images. In addition, we make sure that the views of the under- and over-exposed images are different from the view of the mid-exposed reference image, in order to simulate the required difference in terms of scene content between the reference and non-reference images. For example, if the reference LDR image is set to the left view, the under- and over-exposed images are set to the right view, and vice versa. In addition, we enlarge the training set by including additional scenes from the free-motion dataset provided by Karaduzovic-Hadziabdic \emph{et al.} in~\cite{karaduzovic2016subjective}. This dataset contains several scenes with differently exposed instances of a specific view which we select as the reference view, as well as additional differently exposed LDR images depicting various types of object motion. The combination of the stereo datasets and the free-motion dataset from~\cite{karaduzovic2016subjective} is crucial for the generalization performance of the desired exposure fusion for dynamic scenes model. 
  
  All sub-networks in the case of $3$-LDR based learnable exposure fusion are composed of $3$ convolutional and $3$ deconvolutional layers, including the modifications suggested in~\cite{Ibrahim2016Segm}. Each convolutional layer has $16$ filters. Furthermore, we modify the network architecture of each sub-network by including $2$ additional convolutions after each level in the contractive and refinement parts except for the third convolutional layer corresponding to the lowest resolution, where we add $4$ such convolutions. This modification is suggested in the original \emph{FlowNet}~\cite{Dobrovsky15Flown} design as well as other works~\cite{he2016deep, simo2016learning}. The additional convolutions do not change the spatial resolution of the input data, and guarantee a better feature abstraction. This modification is essential for the $3$-LDR scenarios as the inputs to the \emph{exposures merging} and the subsequent \emph{guided de-ghosting} sub-networks are presented as a tensor involving multiple concatenated images. 
  
\begin{figure*}[h!]
     \includegraphics[width=\textwidth, height = 95mm]{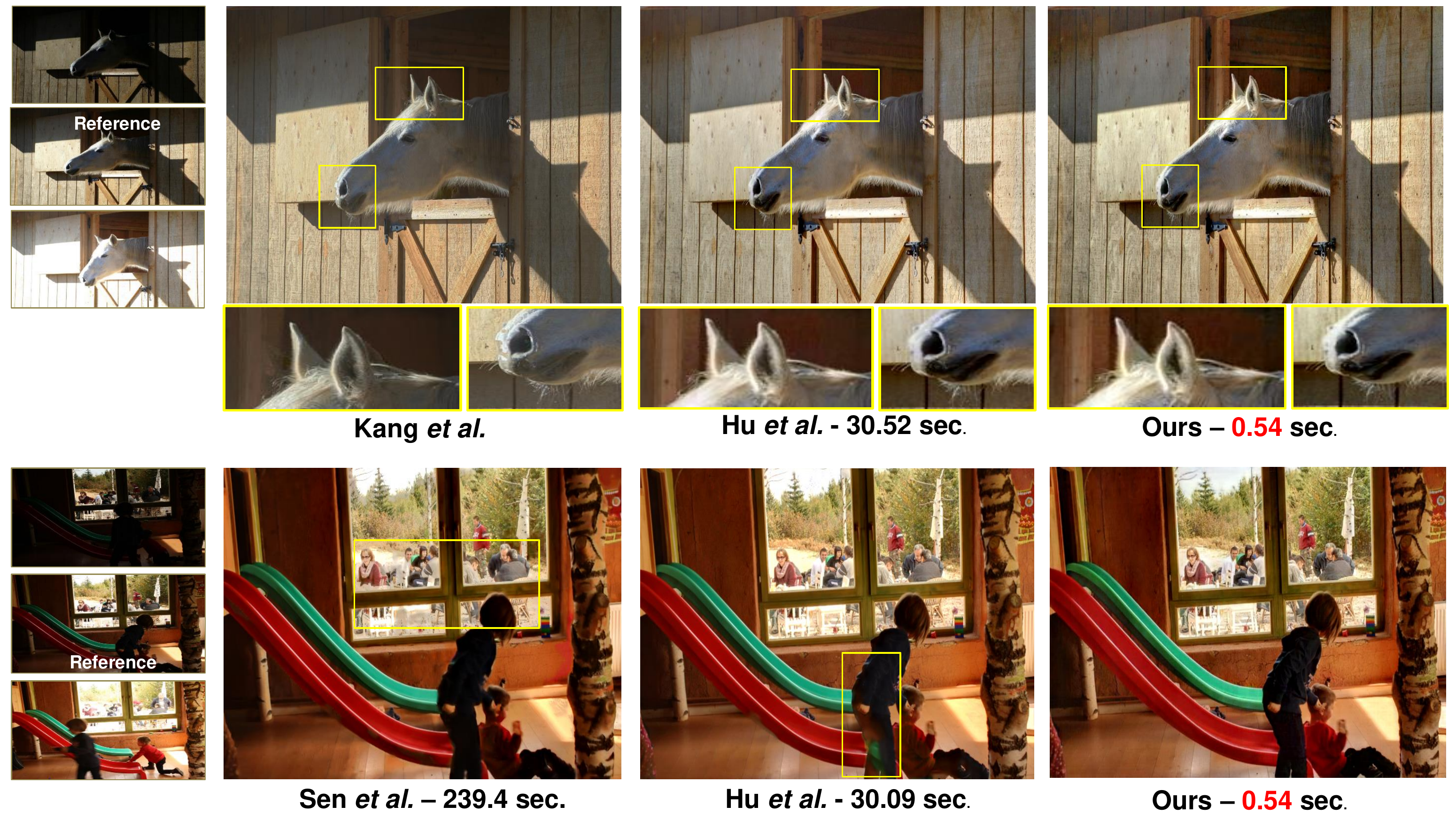}
    \caption{Additional comparison of ours results on scenes composed of $3$ LDR images, against the methods of Kang \emph{et al.}~\cite{kang2003high} (first scene)}, Sen \emph{et al.} (second scene) and Hu \emph{et al.}. In addition we provide the corresponding executions times of our approach as well as the methods of Sen \emph{et al.} and Hu \emph{et al.} (for the alignment part). Our exposure fusion for dynamic scenes model yields artifact-free images of the corresponding reference view, despite the challenging conditions of the scenes in terms of the exposure difference and the amount of motion. This is achieved in a short time span. Images for the first scene (first row) are courtesy of Kang \emph{et al.}~\cite{kang2003high}, while the images of the second scene (second row) are courtesy of Kanita Karaduzovic-Hadziabdic~\cite{expEval14Hdr}.  
     
    \label{fig:Fig10}
\end{figure*}

 Figure~\ref{fig:Fig10} contains a the results of our exposure fusion for dyamnic scenes model for the case of $3$ images, together with the results of the methods of Kang \emph{et al.}~\cite{kang2003high}, Sen \emph{et al.}~\cite{PatchBasedHDR} and \emph{Hu et al.}~\cite{Hu2013Hdrdeg}. Our rendering model is capable of handling several types of motion as well as large exposure difference between the input images. In fact, our approach performs particularly well when dealing with regions that are under- and/or over-exposed in the input images. Whereas the state-of-the-art approaches fail to accurately reconstruct the details in such regions. Together with the high quality output image produced by our model, the computation time required is very low in comparison to other methods.

%
%

\section{Conclusion}

We propose an end-to-end multi-module CNN architecture which learns to perform exposure fusion on input LDR images presenting scene and color differences. A distinctive aspect of our approach is that our images are used as input at multiple different stages of the CNN architecture. We propose solutions for $2$-image and $3$-image LDR input cases, where each case is provided with a different architecture. In various comparisons with state-of-the-art on multiple datasets, we show that the proposed approach yields excellent results. It successfully removes ghost artifacts and maintains high contrast in the obtained output image.

{\small
\bibliographystyle{ieee}
\bibliography{egbib}
}

\end{document}